\documentclass[11pt]{article}
\usepackage[final]{acl}
\usepackage{times}
\usepackage{latexsym}
\usepackage[T1]{fontenc}
\usepackage[utf8]{inputenc}
\usepackage{microtype}
\usepackage{inconsolata}
\usepackage{graphicx}
\usepackage{enumitem}
\usepackage{booktabs}
\usepackage{amssymb}
\usepackage{amsmath}
\usepackage{colortbl}
\usepackage{amsfonts}
\usepackage[table]{xcolor} 
\usepackage{stfloats}
\usepackage{float}
\usepackage{subcaption}

\title{AFMRL: Attribute-Enhanced Fine-Grained Multi-Modal Representation Learning in E-commerce}

\author{
Biao Zhang\thanks{Equal Contribution.}, 
Lixin Chen\footnotemark[1], 
Bin Zhang\footnotemark[1],
Zongwei Wang\thanks{Project Leader.},
Tong Liu\thanks{Corresponding Author.}, 
Bo Zheng \\
\\
Taobao \& Tmall Group of Alibaba\\
}

\begin{document}
\maketitle
\begin{abstract}

Multimodal representation is crucial for E-commerce tasks such as identical product retrieval. Large representation models (e.g., VLM2Vec) demonstrate strong multimodal understanding capabilities, yet they struggle with fine-grained semantic comprehension, which is essential for distinguishing highly similar items. To address this, we propose Attribute-Enhanced Fine-Grained Multi-Modal Representation Learning (AFMRL), which defines product fine-grained understanding as an attribute generation task. It leverages the generative power of Multimodal Large Language Models (MLLMs) to extract key attributes from product images and text, and enhances representation learning through a two-stage training framework: 1) Attribute-Guided Contrastive Learning (AGCL), where the key attributes generated by the MLLM are used in the image-text contrastive learning training process to identify hard samples and filter out noisy false negatives. 2) Retrieval-aware Attribute Reinforcement (RAR), where the improved retrieval performance of the representation model post-attribute integration serves as a reward signal to enhance MLLM's attribute generation during multimodal fine-tuning.  Extensive experiments on large-scale E-commerce datasets demonstrate that our method achieves state-of-the-art performance on multiple downstream retrieval tasks, validating the effectiveness of harnessing generative models to advance fine-grained representation learning.

\end{abstract}

\section{Introduction}

\begin{figure}[t!]
    \centering
    \includegraphics[width=\linewidth]{images/fig1.png}
    \caption{Comparison of multimodal information in General and E-commerce domains. In General domains, text typically provides a global description of the image, with multiple instances corresponding between them. In contrast, in E-commerce domains, product titles often describe only specific instances within the product image (which is called the main subject). Moreover, for the same product, there may exist various images (e.g., different shooting angles and backgrounds) and titles (e.g., varying descriptions, marketing phrases).}
    \label{fig:eco_dom}
\end{figure}

The field of multimodal representation learning is undergoing a paradigm shift, moving beyond discriminative matching frameworks towards generative models capable of sophisticated understanding and reasoning \cite{jiang2025vlm2vec, zhang2025gme, zhang2024notellm, zhou-etal-2025-megapairs}. This evolution is particularly critical in domains like E-commerce, where the ability to distinguish between visually similar products hinges on a deep comprehension of fine-grained attributes. Accurately retrieving a product requires not just matching a query like \textit{"red dress"} to an image, but understanding nuanced details such as \textit{"a V-neck A-line dress in crimson silk with cap sleeves"}, which necessitates a model that can parse compositional structure and subtle visual cues. This paper explores the transition from traditional representation models to Multimodal Large Language Models (MLLMs) to achieve this leap in capability.

Traditional representation models, such as CLIP \cite{radford2021clip}, are built on discriminative dual-encoder architectures that learn an aligned metric space for matching. This approach, while effective for broad semantic retrieval, often functions as a "bag-of-words" system \cite{yuksekgonul2022bag_of_words}, struggling with compositional reasoning. For instance, it can fail to robustly distinguish between "a \textit{white} t-shirt with a \textit{blue} logo" and "a \textit{blue} t-shirt with a \textit{white} logo." In stark contrast, MLLMs, built on auto-regressive principles, are compelled to understand such structural relationships in order to produce coherent descriptions sequentially. This inherent design not only embeds compositional understanding but also unlocks transformative advantages, including instruction-driven flexibility to create task-aware representations and emergent commonsense reasoning to infer higher-level abstract concepts \cite{li2025highlighting}.

Despite its potential, adapting MLLMs to fine-grained representation learning still faces a fundamental challenge. Constrained by the causal attention mechanism, prevailing large representation models typically derive embeddings via global average pooling (e.g., LLM2Vec \cite{behnamghader2024llm2vec}) or last-token hidden states (e.g., VLM2Vec \cite{jiang2025vlm2vec}), which is incompatible with established fine-grained alignment techniques such as Region-of-Interest (RoI)-based methods \cite{xie2025fgclip}. This limitation inspires our core research question: \textbf{How can we harness an MLLM's advanced understanding and instruction-following abilities to overcome its architectural constraints for fine-grained E-commerce tasks?}

Our solution is to define fine-grained product understanding as a key attribute generation task. Continuing with the \textit{“red dress”} as an example, if MLLMs can infer key attributes such as \textit{“dark red”}, \textit{“silk”}, \textit{“V-neck”}, \textit{“A-line”}, and \textit{“cap sleeves”} from the product’s image and text, we consider it to have strong fine-grained understanding capability. The next question we need to address is: \textbf{How can we integrate the generated key product attributes into multimodal representation learning?} To this end, we propose a two-stage training framework: 1) Attribute-Guided Contrastive Learning (AGCL): We first employ an MLLM to generate a list of critical attributes for each product (e.g., "suede material," "lace-up closure," "rubber sole"). These attributes then serve as an explicit supervisory signal within the contrastive learning process, helping identify hard negatives and filter out noisy false negatives, thereby sharpening the model’s discriminative power; 2) Retrieval-aware Attribute Reinforcement (RAR): To resolve the potential misalignment between the generated textual attributes and the learned visual representation, we introduce a reinforcement learning stage. The representation model itself acts as a reward function, providing feedback to fine-tune the attribute generation model. This creates a self-improving loop, ensuring that the generated attributes are maximally aligned with and beneficial for the final discriminative representation.

In summary, this work makes the following contributions:
\begin{itemize}[noitemsep,topsep=0pt]
\item To the best of our knowledge, we systematically investigate and validate the application of Multimodal Large Language Models (MLLMs) for fine-grained representation learning within the E-commerce domain.
\item We propose a novel framework that integrates Attribute-Guided Contrastive Learning (AGCL) and Retrieval-aware Attribute Reinforcement (RAR), significantly improving the model's fine-grained discriminative capability while ensuring alignment between attribute generation and representation learning.
\item Extensive experiments on challenging E-commerce datasets demonstrate that our method achieves state-of-the-art performance across multiple downstream retrieval tasks, validating both its effectiveness and superiority.
\end{itemize}

\section{Related Work}

\subsection{Fine-grained Multimodal Understanding}

The foundation of modern multimodal representation learning was laid by models like CLIP, which pioneered large-scale contrastive learning between images and text. Building on this, a range of approaches have been proposed to enhance fine-grained vision-language understanding. For example, FG-CLIP \cite{xie2025fgclip} improves cross-modal alignment by leveraging hard negative samples and region-level annotations; ECLIP \cite{jin2023eclip} introduces instance-centric pretraining for E-commerce applications, while $\text{E}^2$ \cite{qi2024e2} employs regional contrastive learning for fashion retrieval. These methods typically enhance model discriminability by incorporating more granular signals.

In recent years, Multimodal Large Language Models (MLLMs) have emerged as powerful tools for multimodal understanding. Models such as LLaVA \cite{liu2023llava}, CogVLM \cite{wang2024cogvlm}, DeepSeek-VL  \cite{lu2024deepseek}, and Qwen2.5-VL \cite{bai2025qwen2_5} leverage the world knowledge and reasoning capabilities of large language models to effectively extract detailed information from both images and text, achieving significantly stronger performance in fine-grained multimodal tasks.

\subsection{Generative Models for Representation Learning}

The remarkable success of Large Language Models (LLMs) has motivated a recent surge in research aimed at repurposing decoder-only architectures for dense representation learning \cite{behnamghader2024llm2vec, lee2024nv_embed, ma2024ft_llama_retrieve, shin2025generative_prob_repre}. Early efforts, such as LLM-Embedder \cite{zhang2023llm_embedder}, adapted prompt-based techniques to extract representations. More recent approaches have introduced architectural modifications; for instance, LLM2Vec \cite{behnamghader2024llm2vec} enables bidirectional attention and employs a masked next-token prediction objective to convert pre-trained decoders into powerful text encoders. Similarly, NV-Embed \cite{lee2024nv_embed} improves embedding efficiency by incorporating a latent attention mechanism and removing the causal mask during contrastive training.

This trend has naturally extended into the multimodal domain, leveraging the superior architectural and reasoning capabilities of MLLMs. Models such as MagicLens \cite{zhang2024magiclens} have demonstrated strong performance in instruction-guided retrieval, while frameworks like E5-V \cite{jiang2024e5v} and VLM2Vec \cite{jiang2025vlm2vec} have set new benchmarks for encoding long, complex multimodal inputs into universal embeddings.

\begin{figure*}[h]
    \centering
    \includegraphics[width=1.0\textwidth]{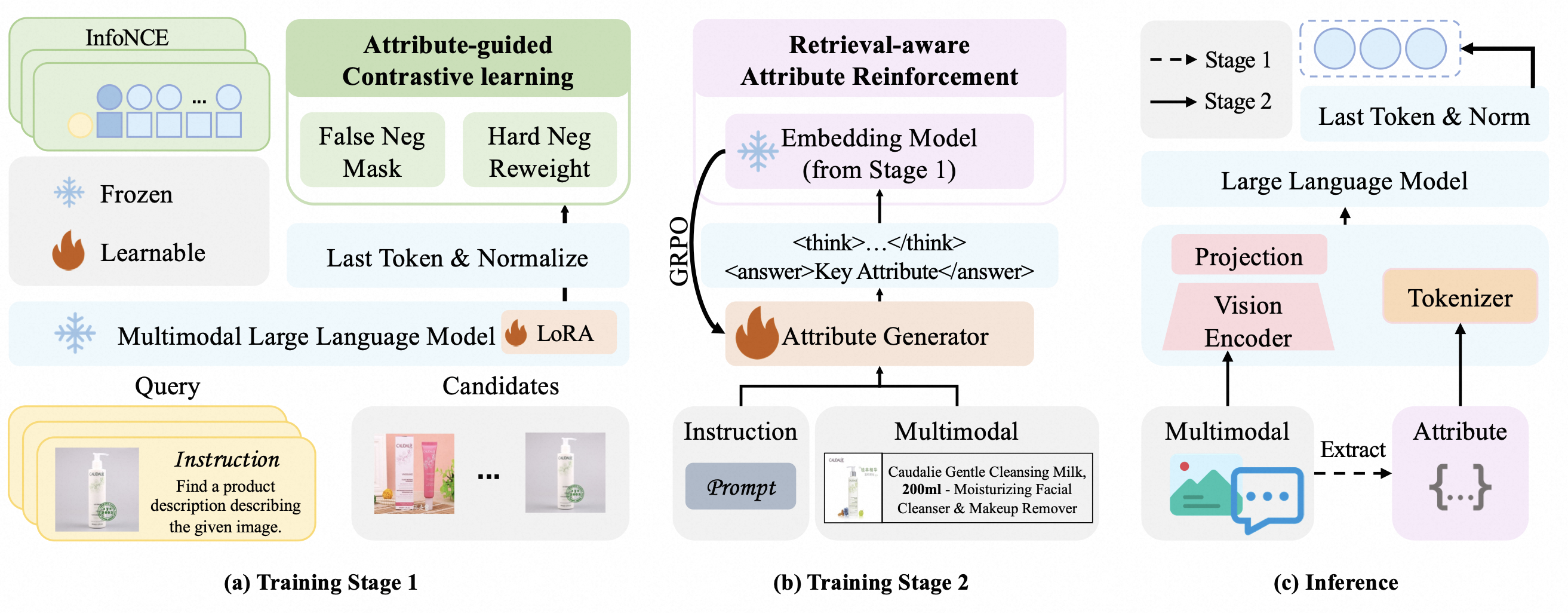}
    \caption{An overview of our proposed framework. The model is trained in two stages. (a) Stage 1: Attribute-Guided Contrastive Learning: A representation model is trained using an enhanced contrastive loss with false negative masking and hard negative reweighting. (b) Stage 2: Retrieval-aware Attribute Reinforcement: An attribute generator is fine-tuned with reinforcement learning (GRPO), using the frozen representation model to provide a direct, retrieval-based reward. (c) Inference: The optimized generator extracts attributes to enrich the query input, which is then encoded by the representation model for retrieval.}
    \label{fig:main_frame}
\end{figure*}

\section{Methodology}

In this section, we present AFMRL (Attribute-Enhanced Fine-Grained Multi-Modal Representation Learning), a novel framework for learning fine-grained multimodal representations. Our AFMRL framework is designed around a principle of decoupled responsibilities. We employ two specialized models: a highly efficient Representation Model optimized for generating discriminative embeddings, and a powerful Attribute Generator dedicated to high-level reasoning and extracting key local features. This separation allows each component to excel at its specific task. The central premise of our work is that explicitly incorporating these attributes can resolve critical ambiguities in fine-grained retrieval, particularly with hard negatives---items that are visually similar but semantically distinct.

\begin{figure}[ht]
    \centering
    \includegraphics[width=0.9\linewidth]{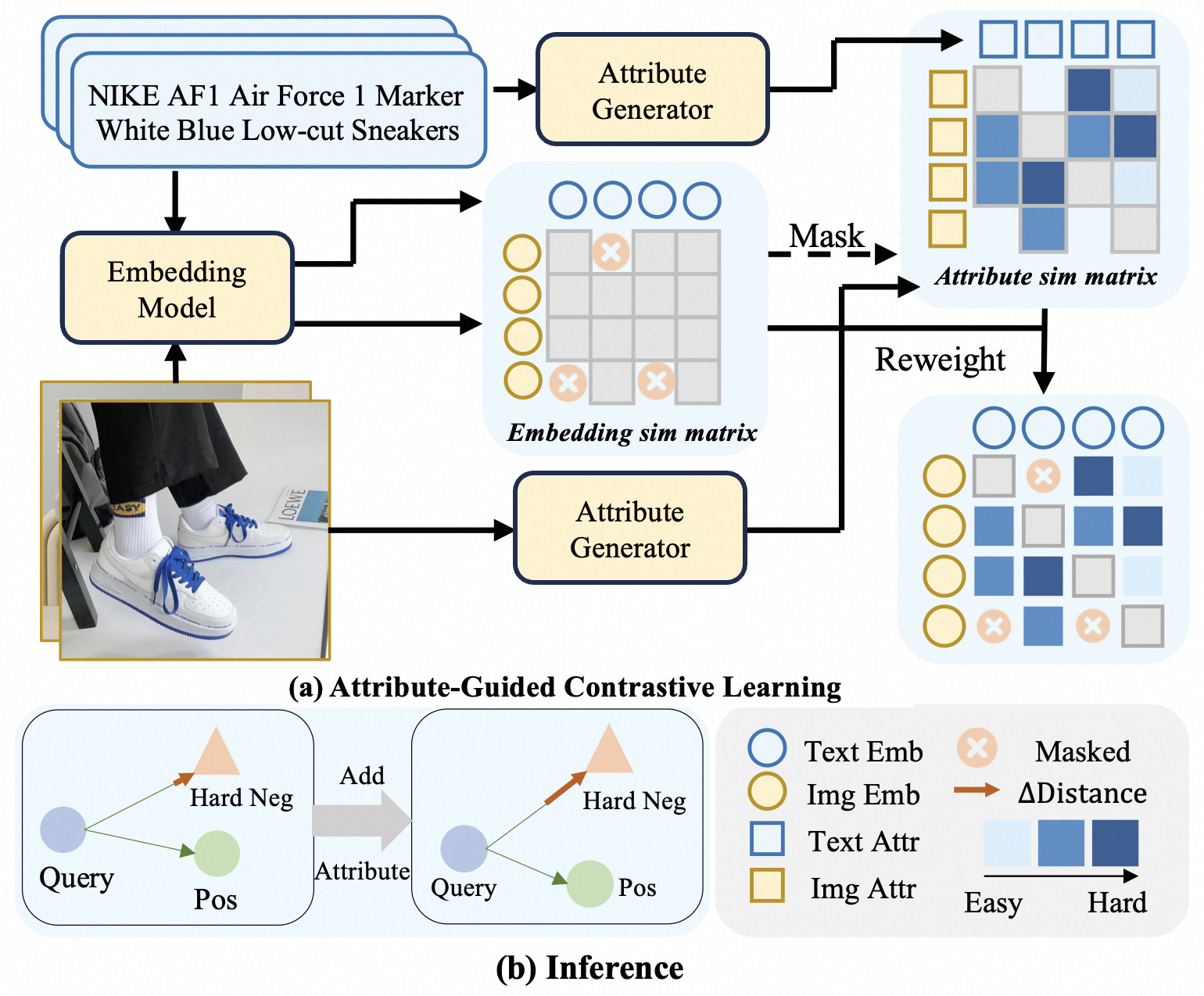}
    \caption{Demonstration of our proposed Attribute-Guided Learning process and attribute's influence on the retrieval.}
\label{fig:method_motivation}
\end{figure}

\subsection{Attribute-Guided Contrastive Learning}
\label{sec:agcl}

\begin{figure*}[ht]
    \centering
    \includegraphics[width=1.0\linewidth]{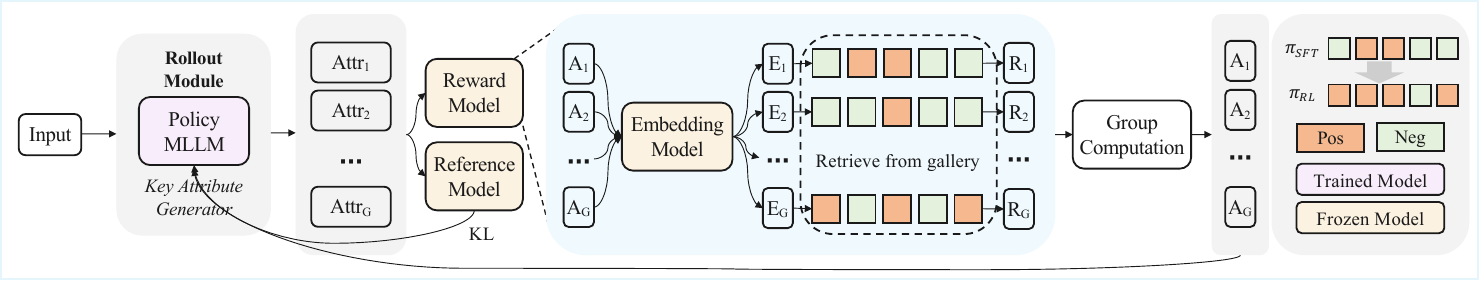}
    \caption{The overall framework of our proposed Retrieval-aware Attribute Reinforcement training pipeline. G denotes the Attribute Generator, and R denotes the Representation Model (the frozen encoder used for reward calculation).}
    \label{fig:method_rl}
\end{figure*}

As illustrated in Figure \ref{fig:method_motivation}, AGCL utilizes VLM2Vec \cite{jiang2025vlm2vec} as the base embedding model. Following CLIP, VLM2Vec is trained with contrastive learning using the InfoNCE loss. This leads to two key issues: (1) failing to leverage complementary matching signals beyond the dense embeddings, and (2) penalizing the model for matching with false negatives (semantically similar items in the batch).

To address these challenges, AGCL introduces an MLLM-based attribute generator (distilled from Qwen2.5-VL-72B-Instruct \cite{bai2025qwen2_5}; details in Appendix \ref{sec:generator}) to produce key attributes for each training sample, and enhances the standard InfoNCE loss. Specifically, we utilize the generated key attributes to guide the selection of hard negatives. We compute BM25 scores, $B_{ij}$, between the key attributes of query $q_i$ and each candidate $p_j$ to quantify their lexical relevance. BM25 is a robust information-retrieval function. High BM25 scores identify samples that are lexically similar to the query, making them challenging negatives that warrant greater attention during training.
To integrate these scores, we transform them into normalized importance weights using a bounded activation function:
\begin{equation}
w_{ij} = e^{1 + \tanh( B_{ij})}.
\end{equation}
This formulation ensures weights are bounded, providing stable and targeted emphasis on lexically hard negatives. 
In addition, for each query $q_i$, we compute cosine similarities $s_{ij} = \cos(q_i, p_j)$ with all candidate samples $p_j$ in the batch, and then define a binary masking function:
\begin{equation}
    \mathcal{M}_{ij} = \mathbb{I}[(j \neq i) \land (s_{ij} > s_{ii} + \delta)],
\end{equation}
where $\mathbb{I}[\cdot]$ is the indicator function. If a candidate negative sample’s similarity to the query exceeds a predefined margin, we remove it from the in-batch pool. This creates a refined set of valid negatives $\mathcal{N}_i = \{j : \mathcal{M}_{ij} = 0\}$ for each query, ensuring the model is not penalized for correctly identifying semantically similar samples.

The final loss function holistically integrates the two strategies above. For each query $q_i$ with similarity scores scaled by temperature $\tau$, the AGCL loss is formulated as:
\begin{equation}
\mathcal{L}_{\text{AGCL}} = -\log \frac{w_{ii}\cdot e^{s_{ii}/\tau}}{w_{ii}\cdot e^{s_{ii}/\tau} + \sum_{j \in \mathcal{N}_i} w_{ij} \cdot e^{s_{ij}/\tau}}.
\end{equation}
Here, the summation in the denominator is restricted to the refined negative set $\mathcal{N}_i$, thereby excluding false negatives. The remaining valid negatives are re-weighted by their importance scores $w_{ij}$. This integrated objective enables the model to learn more robust and discriminative representations by focusing on true, informative negative samples.

\subsection{Retrieval-aware Attribute Reinforcement}

Section \ref{sec:agcl} yields a strong foundation: a Key Attribute Generator ($\pi_{\text{SFT}}$) proficient at producing plausible attributes, and a powerful Embedding Model trained with AGCL. However, the generator's distillation objective is disconnected from the final retrieval task. To bridge this optimization gap, we introduce a reinforcement learning (RL) stage that directly aligns the attribute generation policy with downstream retrieval performance. This RL stage fine-tunes the generator policy, $\pi_{\theta}$, to produce attributes that maximize the efficacy of the fixed representation model.

\textbf{Direct-Feedback Reward Mechanism.}
The design of the reward function is paramount to the effectiveness of reinforcement learning. Moving beyond simplistic proxy signals, we introduce a reward function that is directly coupled with our final task metric. We accomplish this by leveraging our pre-trained representation model as an integral component of the reward evaluation environment. In this setup, the policy $\pi_{\theta}$ generates a set of attributes for a given query. These attributes are used to augment the multimodal input, which is subsequently passed to the representation model to conduct a retrieval search across a candidate pool. The reward is then defined precisely as the Recall@k of this search.

\begin{table*}[h]
\centering
\resizebox{2\columnwidth}{!}{
\begin{tabular}{lccccccccccc}
\toprule
& \multicolumn{3}{c}{Image-to-Text Retrieval} & & \multicolumn{3}{c}{Text-to-Image Retrieval} & & \multicolumn{3}{c}{Coarse-grained Product Retrieval} \\ 
\cmidrule(lr){2-4} \cmidrule(lr){6-8} \cmidrule(l){10-12} 
Model & Recall@1 & Recall@5 & Recall@10 & & Recall@1 & Recall@5 & Recall@10 & & mAP@1 & mAP@5 & mAP@10 \\ 
\midrule
\multicolumn{12}{l}{\textit{BERT-based Baselines}} \\
CLIP & 21.5 & 43.6 & 54.6 & & 19.5 & 40.9 & 52.1 & & 61.7 & 63.6 & 63.1 \\
ALBEF & 22.4 & 46.1 & 57.4 & & 20.2 & 43.2 & 54.9 & & 63.2 & 65.4 & 64.7 \\ 
BLIP & 24.9 & 48.5 & 59.3 & & 23.5 & 46.2 & 57.5 & & 64.0 & 65.9 & 65.1 \\ 
Region-CLIP & 21.6 & 43.8 & 55.1 & & 19.2 & 40.3 & 51.7 & & 62.0 & 63.9 & 63.9 \\ 
FG-CLIP & 21.8 & 43.9 & 55.3 & & 19.4 & 40.2 & 51.5 & & 63.3 & 65.9 & 64.1 \\ 
VL-CLIP & 21.5 & 43.8 & 54.9 & & 19.2 & 39.9 & 50.5 & & 62.9 & 64.9 & 63.4 \\ 
\midrule
\multicolumn{12}{l}{\textit{LLM-based Baselines}} \\
LLM2CLIP & 30.1 & 56.8 & 68.2 & & 27.7 & 54.3 & 66.2 & & 68.3 & 71.9 & 70.4 \\
+ AGCL & 32.0& 58.6 & 69.7 & & 29.5 & 55.9 & 67.5 & & 69.1 & 72.9 & 71.3 \\
VLM2Vec & 30.5 & 55.2 & 66.1 & & 36.4 & 63.0 & 73.6 & & 70.3 & 76.9 & 72.1 \\
+ AGCL & 32.8 & 57.0 & 67.7 & & 38.1 & 64.5 & 75.0 & & 71.3 & 78.3 & 73.2 \\ 
\bottomrule
\end{tabular}
}

\caption{Performance on Coarse-Grained and Cross-Modal Retrieval Tasks. The AGCL component of our AFMRL framework provides consistent improvements over strong baselines, demonstrating its broad effectiveness for general retrieval.}
\label{tab:main_res}
\end{table*}

\begin{table*}[ht]
\centering
\resizebox{1.6\columnwidth}{!}{
\begin{tabular}{lcccccc}
\toprule
Model                 & Recall@1       & Recall@5       & Recall@10      & \multicolumn{1}{l}{NDCG@1} & \multicolumn{1}{l}{NDCG@5} & \multicolumn{1}{l}{NDCG@10} \\ \midrule
\multicolumn{4}{l}{\textit{BERT-based Baselines}}                        & \multicolumn{1}{l}{}       & \multicolumn{1}{l}{}       & \multicolumn{1}{l}{}        \\
CLIP              & 14.98 & 23.07 & 27.59 & 14.98 & 19.20 & 20.66 \\
ALBEF             & 20.59 & 30.13 & 34.46 & 20.59 & 25.65 & 27.05 \\ 
BLIP              & 19.42 & 29.05 & 33.27 & 19.42 & 24.45 & 25.81 \\ 
Region-CLIP       & 29.35 & 43.78 & 58.49 & 30.01 & 39.98 & 46.86 \\ 
FG-CLIP           & 31.44 & 49.78 & 68.38 & 28.77 & 45.29 & 50.81 \\ 
VL-CLIP           & 30.49 & 48.98 & 64.57 & 26.83 & 41.99 & 49.31 \\ \midrule
\multicolumn{4}{l}{\textit{LLM-based Baselines \& Ablations}}            & \multicolumn{1}{l}{}       & \multicolumn{1}{l}{}       & \multicolumn{1}{l}{}        \\
VLM2Vec           & 48.05 & 64.26 & 69.65 & 48.58 & 59.00 & 60.90 \\
+ AGCL            & 51.06 & 68.08 & 73.52 & 51.06 & 60.19 & 62.02 \\
+ AGCL + Distill Gen. & 52.42 & 71.00 & 76.26 & 52.42 & 62.64 & 64.34 \\
\textbf{AFMRL (Ours)} & \textbf{54.28} & \textbf{72.19} & \textbf{77.27} & \textbf{54.28}  & \textbf{64.08} & \textbf{65.69}  \\ \bottomrule
\end{tabular}
}
\caption{Fine-Grained Instance Retrieval Results. Our full model, AFMRL, demonstrates state-of-the-art performance, with each component providing a clear incremental benefit.}
\label{tab:fine_grained_product_retrieval}
\end{table*}

This direct-feedback loop ensures the policy is optimized precisely for enhancing top-$k$ retrieval. The reward $\mathcal{R}(x,a)$ for a query $x$ and generated attributes $a$ is:
\begin{equation}
    \mathcal{R}(x, a) = \frac{|\mathcal{P}_x \cap \text{top}_k(x, a)|}{|\mathcal{P}_x|},
\end{equation}
where $\mathcal{P}_x$ is the set of ground-truth positives, and $\text{top}_k(x, a)$ denotes the set of top-$k$ retrieved items 
when using generated attributes $a$ for query $x$. To maintain linguistic coherence learned during SFT, we penalize malformed generations. The final reward is:
\begin{equation}
    \mathcal{R}(x, a, \phi) = \mathcal{R}(x, a) \cdot \mathbb{I}[\phi=1] + \eta \cdot \mathbb{I}[\phi=0],
\end{equation}
where $\phi$ is a validity flag and $\eta = -0.1$ is a penalty for invalid outputs.

\textbf{Policy Optimization with GRPO.}
Equipped with the SFT-initialized policy $\pi_{\text{SFT}}$ and our direct-feedback reward $\mathcal{R}$, we employ GRPO  \cite{shao2024grpo} for policy optimization. GRPO is well-suited for this task due to its stability and sample efficiency, mitigating risks associated with potentially sparse rewards. Moreover, it is an efficient algorithm that eliminates the need for an explicit reward model and value models. The RL process fine-tunes the policy $\pi_{\theta}$ (initialized from $\pi_{\text{SFT}}$) to maximize the expected retrieval reward. The objective function of GRPO is defined as follows:

\begin{equation}
\begin{split}
\mathcal{J}_{\text{GRPO}}(\theta) = \mathbb{E}_{\tau \sim \pi_{\theta_{\text{old}}}} \bigg[ &\min \Big( r_t(\theta) \hat{A}_t, \\
&\quad \operatorname{clip}\big(r_t(\theta), 1-\epsilon, 1+\epsilon\big)\\ \hat{A}_t \Big) 
&- \beta D_{\text{KL}}(\pi_{\theta} || \pi_{\text{SFT}}) \bigg],
\end{split}
\label{eq:grpo_objective}
\end{equation}
where $r_t(\theta)$ is the probability ratio, and the advantage estimate $\hat{A}_t$ is computed by normalizing batch rewards: $\hat{A}_i = \frac{(\mathcal{R}_i - \text{mean}(\mathcal{R}))}{\text{std}(\mathcal{R})}$. The KL-divergence term, regulated by $\beta$, acts as a regularization mechanism, anchoring the policy to the robust linguistic foundation established during SFT while encouraging exploration towards higher retrieval performance. $\epsilon$ denotes the clipping parameter in the GRPO objective function, used to prevent excessive policy updates.

\section{Experiments}
We conduct a series of experiments to evaluate AFMRL on large-scale E-commerce benchmarks and ablate the contributions of its key components.

\subsection{Experimental Setup}

\textbf{Dataset.}
We evaluate our method on the large-scale M5Product E-commerce dataset \cite{dong2022m5product}. To ensure data quality, we filter out items with missing or invalid images, resulting in a clean set of 5,760,482 products spanning over 6,000 categories. Additionally, to create more diverse real-world scenarios, we collected a large-scale multimodal product dataset from a popular E-commerce platform, named EIPM (E-commerce Identical Product Matching Dataset). It contains about 2 million groups of same products and over 10 million product items, covering more than 10,000 sub-categories, such as clothes, cosmetics, toys, and so on.

\begin{table}[ht]
\centering

\resizebox{\columnwidth}{!}{
\begin{tabular}{cccccc}
\toprule
AGCL                      & Key Attribute & Accuracy & NMI   & ARI   & Purity \\ \midrule
                          &               & 87.67    & 87.04 & 44.39 & 73.16  \\
\checkmark &              & 87.80    & 87.11 & 44.44 & 73.24  \\

\checkmark & $\pi_{Distill}$  & 87.98    & 87.63 & 46.24 & 74.21  \\ 
\checkmark & $\pi_{RL}$ & \textbf{88.00} & \textbf{87.68} & \textbf{46.61} & \textbf{74.52} \\
\bottomrule
\end{tabular}
}
\caption{Performance on Classification and Clustering Tasks.}
\label{tab:downstream_task}
\end{table}

\textbf{Implementation Details.}
Our AFMRL framework consists of two core components: 1) VLM2Vec Model: We initialize the model from Qwen2-VL-2B-Instruct \cite{wang2024qwen2vl} and fine-tune it using LoRA. The learning rate is 5e-5, the InfoNCE temperature is 0.02, and the batch size is 2048. We use a LoRA rank of 8 and train for 2,000 steps; 2) Attribute Generator: We initialize the generator from Qwen2.5-VL-3B-Instruct \cite{bai2025qwen2_5} and perform full-parameter fine-tuning. The distillation stage uses a learning rate of 5e-5 for 2,000 steps. The margin for the false negative masking strategy is set to 0.4. The RL stage uses GRPO for alignment with a learning rate of 1e-6 for 350 steps. The hyperparameters are set as follows: $\beta = 0.01$ and $\epsilon = 0.2$. The number of rollouts is set to 8. Details of the RL training process are provided in Appendix \ref{sec:train_detail}.

\textbf{Baselines.}
We compare AFMRL against two categories of models: established BERT-based models (CLIP \cite{radford2021clip}, ALBEF \cite{li2021albef}, etc.) and LLM-based approaches (LLM2CLIP \cite{huang2024llm2clip}, VLM2Vec \cite{jiang2025vlm2vec}). This enables a comprehensive evaluation against both classic and state-of-the-art architectures.

\subsection{Main Results}

Depending on the retrieval granularity, product retrieval can be divided into coarse-grained and fine-grained tasks. In coarse-grained retrieval, products belonging to the same category are treated as positive samples, whereas fine-grained retrieval imposes stricter requirements: items are considered positive samples only when attributes (e.g., style and color) match exactly.

\subsubsection{Coarse-Grained and Cross-Modal Retrieval}

As shown in Table \ref{tab:main_res}, compared with traditional discriminative models, representation models based on MLLMs exhibit significantly better retrieval performance. Furthermore, after incorporating AGCL, the model performance is further improved. This indicates that the AGCL serves as a strong foundation for downstream retrieval tasks: even without a dedicated attribute generator, it can still improve the quality of representations for broad semantic matching, thereby providing a more powerful base model for various retrieval tasks.

\subsubsection{Fine-Grained Instance Retrieval}
Fine-grained instance retrieval imposes high demands on discriminative capability. Here, we deploy the full AFMRL framework, including the attribute generator.

As shown in Table \ref{tab:fine_grained_product_retrieval}, models such as Region-CLIP and FG-CLIP, which incorporate local features, achieve much better performance than on coarse-grained retrieval tasks, confirming that fine-grained features are crucial for E-commerce tasks. In contrast, the inability to fully exploit local features nearly erases the representational advantage brought by the strong reasoning capabilities of MLLMs. 

Notably, building on AGCL, the full AFMRL framework achieves state-of-the-art performance: 1) The distilled generator significantly boosts Recall@1 to 52.42\% by adding high-quality descriptive attributes, indicating that key attributes effectively enhance the MLLM’s understanding of fine-grained details; 2) Building on this, we further introduce a retrieval-aware policy, culminating in a final Recall@1 of 54.28\%. This validates our core hypothesis: for challenging fine-grained tasks, explicitly generating and optimizing descriptive attributes based on the end-task retrieval objective is maximally effective.

\subsection{\textbf{Evaluation on Downstream Tasks}}

To assess the generalizability of the learned embeddings beyond retrieval, we evaluated them on downstream tasks of product classification and clustering. On a dataset of 849,207 items across 5,146 classes, we trained a linear probe on frozen embeddings for classification (Accuracy) and applied k-Means for clustering. The results in Table~\ref{tab:downstream_task} show that the model equipped with both AGCL and the RAR strategy achieves the best performance on classification and clustering, indicating that AFMRL exhibits strong generalization capability in E-commerce scenarios.

\section{Discussions}

In this section, we delve deeper into the mechanisms behind our experimental results. We analyze the efficacy of our method's key components, provide insights into the intriguing phenomena observed during reinforcement learning, and candidly discuss the limitations of our current approach.

\begin{figure}[ht]
    \centering
    \includegraphics[width=0.95\linewidth]{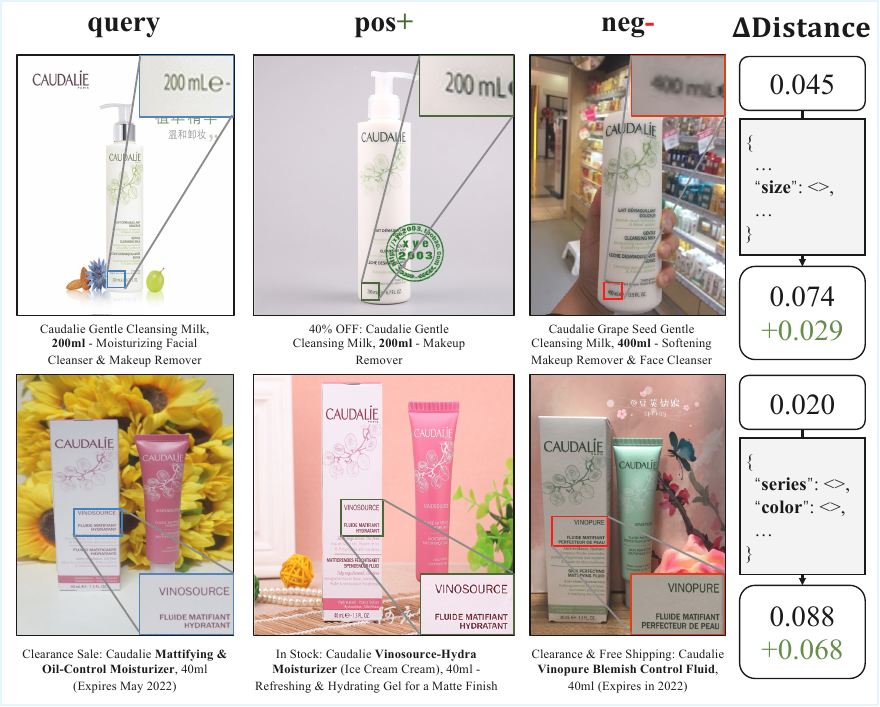}
    \caption{Showcase of key entities.}
    \label{fig:case_entities}
\end{figure}

\subsection{Efficacy of Generated Attributes and the Representation Model}

Our method's core contribution is the use of reinforcement learning to optimize the generation of key attributes that enrich multimodal inputs for retrieval. The effectiveness of this design is twofold: the direct value of the generated attributes and the robust representation model that serves as its foundation.

\textbf{Generated attributes provide more discriminative power.} We argue that the performance gains are primarily attributable to the policy network, $\pi_{\theta}$, generating highly discriminative key attributes. This is intuitively demonstrated by the case study in Figure \ref{fig:case_entities}. In the original embedding space, a query can be equidistant to its positive (pos+) and negative (neg-) samples due to high visual and semantic similarity (e.g., products from the same brand). After augmenting the input with key attributes generated by our model (e.g., "size": "<>", "series": "<>"), the distance to the negative sample is significantly increased ($\Delta \text{Distance}$ grows) while similarity to the positive sample is maintained. This indicates that the generated attributes guide the model to focus on subtle yet critical differences, effectively "pushing away" incorrect candidates and making the feature representations more discriminative.

\begin{figure}[ht]
    \centering
    \includegraphics[width=1.0\linewidth]{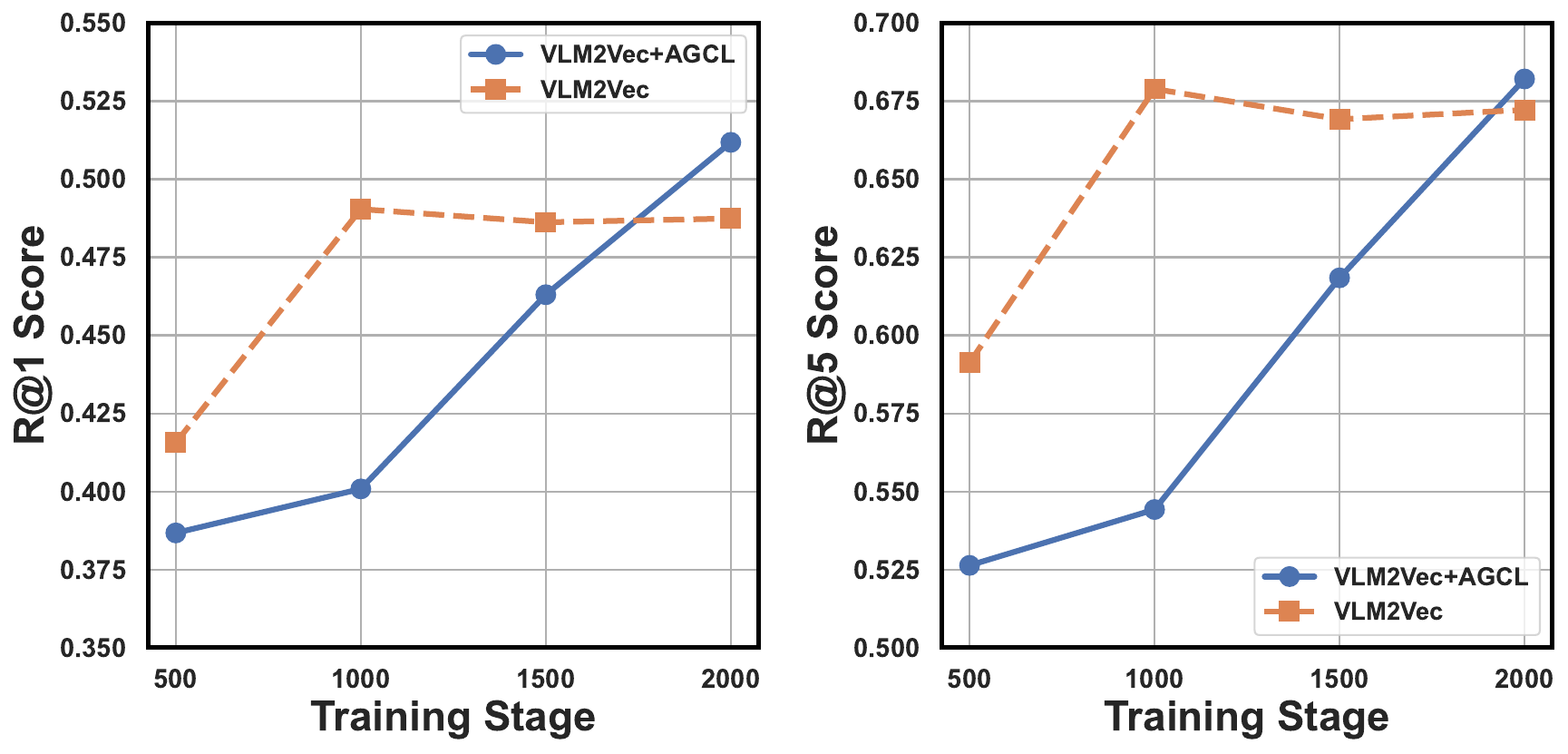}
    \caption{Curve chart (w/o AGCL) of training steps and R@1.}
    \label{fig:curve_chart}
\end{figure}

\textbf{AGCL provides a robust foundation for attribute generation.} The generation of high-quality attributes relies on a powerful underlying representation model. As shown in Figure \ref{fig:curve_chart}, our proposed Attribute-Guided Contrastive Learning (AGCL) strategy plays a pivotal role. Compared to the baseline {\color{black}VLM2Vec}, which quickly plateaus and gets stuck in a local optimum, our proposed component {\color{black}AGCL} exhibits steady and continuous improvement on both R@1 and R@5 metrics, eventually surpassing the baseline. We posit that AGCL encourages the model to learn a more robust and generalized representation space by preventing it from overfitting to simple negatives early in training. This high-quality representation space provides a solid foundation upon which the RL policy can learn to generate precise and effective attributes.

\begin{figure}[ht]
    \centering
    \includegraphics[width=1.0\linewidth]{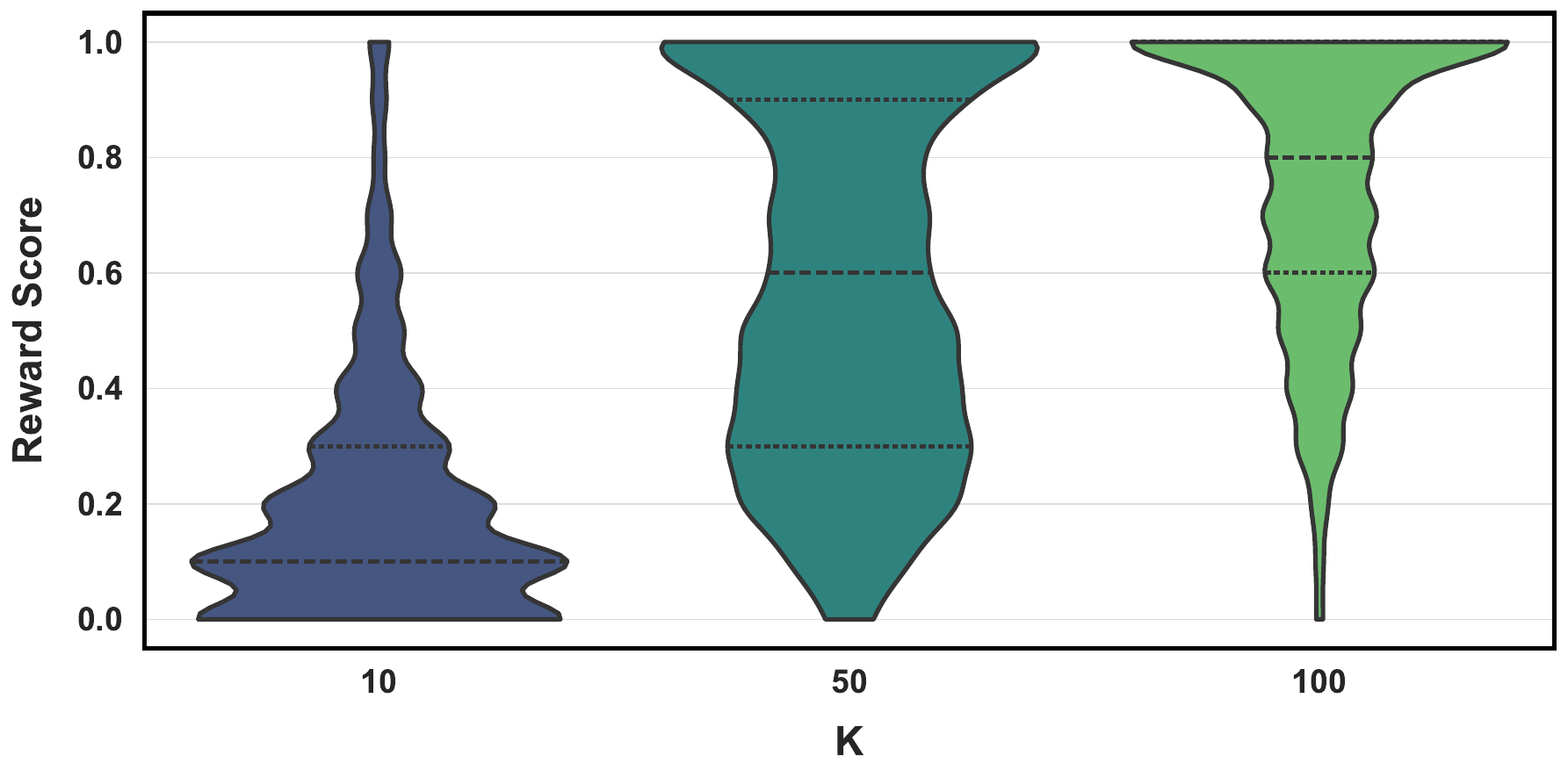}
    \caption{Distribution of reward scores for different $k$.}
    \label{fig:reward_k}
\end{figure}

\subsection{In-depth Analysis of the Reinforcement Learning Process}

Having established the effectiveness of our core design, we now analyze the RL training process itself to understand its internal dynamics and emergent properties.

\textbf{Motivation for the Choice of \(k\).} We use recall as the reward function for RAR. In the Recall@k reward, the hyperparameter \(k\) controls the trade-off between the precision and the density of the learning signal: if \(k\) is too small, the reward becomes overly sparse; if \(k\) is too large, the signal saturates and the gradients weaken. Since GRPO relies on discriminative inter-group reward differences, a well-balanced reward distribution is crucial. As shown in Fig. \ref{fig:reward_k}, \(k=50\) strikes the best balance, avoiding the sparsity of \(k=10\) and the saturation of \(k=100\), thereby providing the richest learning signal. In addition, we also compare how different values of \(k\) affect model convergence under different reward functions; details are provided in Appendix \ref{sec:train_detail}.

\begin{figure}[ht]
    \centering
    \includegraphics[width=1.0\linewidth]{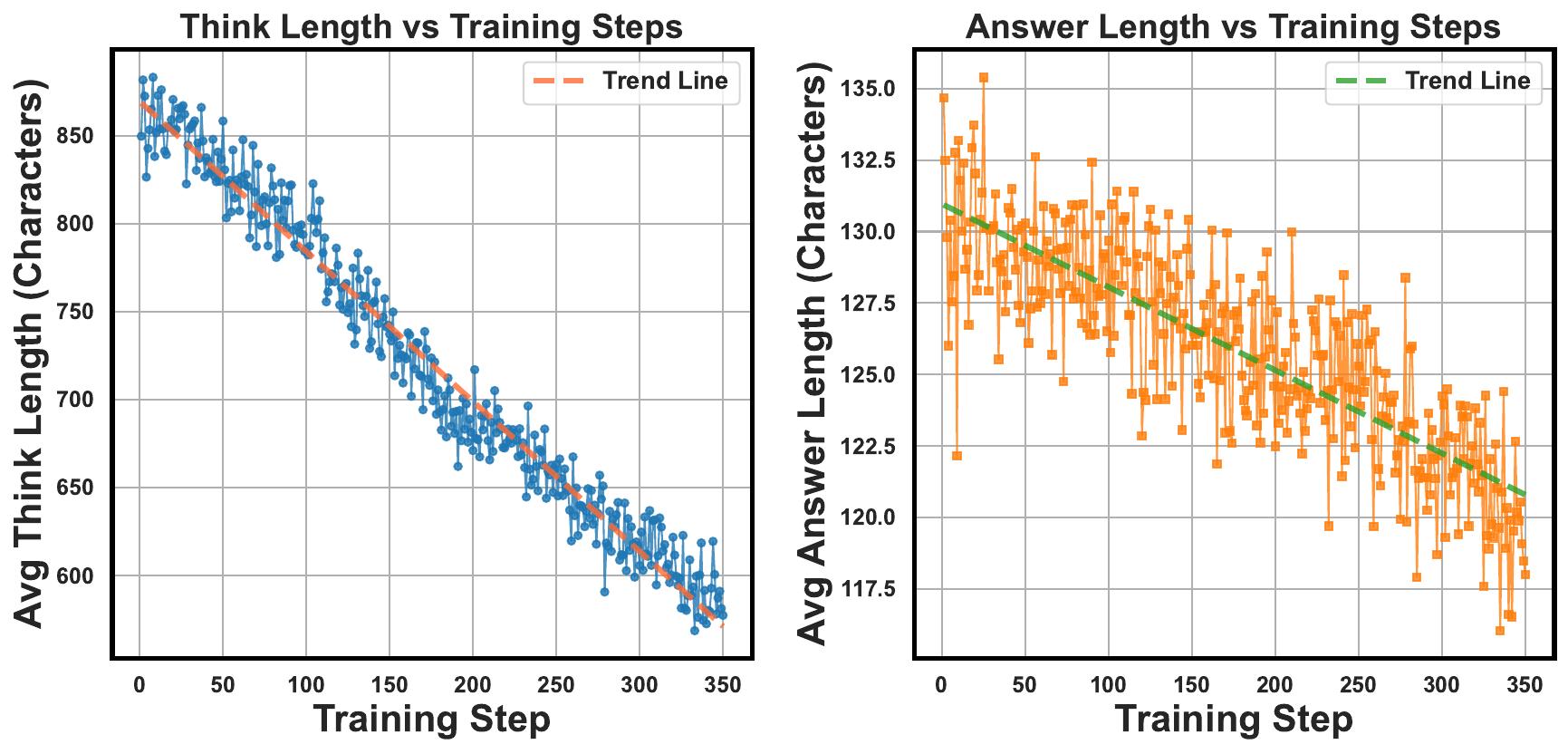}
    \caption{Evolution of think and answer lengths during RL.}
    \label{fig:res_len_rl}
\end{figure}

\textbf{Emergent behavior: generation conciseness.} As illustrated in Figure \ref{fig:res_len_rl}, the average length of generated attributes consistently decreases. Unlike in complex reasoning tasks (e.g., math \cite{shao2024grpo}), where models benefit from longer reasoning chains, our retrieval objective does not require lengthy explanations. The agent learns that redundant or irrelevant attributes act as noise and hurt Recall@k, so it is implicitly encouraged to produce the shortest set of attributes that is still sufficient for good retrieval, leading to concise yet effective generations.

\textbf{Circular Iterative Training.} Our AFMRL framework decouples attribute generation from representation learning, offering an inherent advantage: once the attribute generator is refined via reinforcement learning, it can feed back into the representation learning process to enhance it. We refer to this mechanism as Circular Iterative Training (CIT). As shown in Table~\ref{tab:cit_results}, when we reused the RL-trained attribute generator for AGCL training (using only 30\% of the training samples), the model's performance on downstream tasks improved significantly.

\begin{table}[ht]
  \centering
\resizebox{\columnwidth}{!}{
  \begin{tabular}{lcccc}
    \toprule
    Method & Accuracy & NMI & ARI & Purity \\
    \midrule
    SFT         & 87.80 & 87.11 & 44.44 & 73.24 \\
    SFT+RL      & 88.00 & 87.68 & 46.61 & 74.52 \\
    SFT+RL+CIT  & 89.13 & 88.97 & 47.40 & 75.98 \\
    \bottomrule
  \end{tabular}
}
\caption{Performance comparison of different training strategies on downstream tasks.}
\label{tab:cit_results}
\end{table}

\section{Conclusions}

This paper focuses on a key challenge in fine-grained E-commerce retrieval: representation models based on MLLMs lack sufficient understanding of fine-grained information. To address this, we propose the AFMRL framework, which first uses MLLM-generated attributes to mine hard negative samples, and then performs representation reinforcement learning with retrieval performance as the reward. Extensive experiments on large-scale E-commerce datasets demonstrate that our method achieves state-of-the-art performance, validating our central thesis that guiding discriminative learning with generative reasoning is an effective approach to fine-grained multimodal retrieval.

\section{Limitations}

Despite the strong performance on our primary retrieval task, we acknowledge the method's limitations. Generalization on downstream tasks. As shown in Table \ref{tab:downstream_task}, when evaluated on downstream classification and clustering tasks, our RL-trained policy ($\pi_{RL}$) is slightly higher than the SFT-trained policy ($\pi_{SFT}$). We attribute this to a phenomenon known as the "alignment tax." Because the $\pi_{RL}$ policy is optimized for the highly specific Recall@k retrieval metric, its representations become specialized for this task, potentially at the cost of their generality. SFT, being a more general-purpose tuning paradigm, appears to better preserve the universal features required for broader tasks like classification and clustering.

\bibliography{ref}

\clearpage
\appendix

\section{Evaluation on EIPM dataset}

To further validate the robustness and generalizability of our proposed method, we conducted additional experiments on a large-scale real-world E-commerce dataset.

\textbf{Setup.} 
It contains 10 million products for training and a test set of 200,000 products. The evaluation tasks, metrics, and baseline models used here are identical to the experimental setup described in the main body of the paper.

\textbf{Results.}
The results of this supplementary experiment are presented in Table \ref{tab:main_online_res}. The key takeaway is that the performance trends observed on this large-scale dataset are consistent with our findings from the primary experiments.

Most importantly, our proposed AGCL method remains highly effective. As shown in Table \ref{tab:main_online_res}, applying AGCL to both the LLM2CLIP and the stronger VLM2Vec backbones results in clear and consistent performance gains across all three retrieval tasks (I2T, T2I, and Coarse-grained). For example, 
$\text{VLM2Vec}_{\text{$AGCL$}}$ outperforms the vanilla VLM2Vec by a significant margin, achieving the best results on all metrics.

This experiment confirms that our approach is not only effective on standard academic benchmarks but also robust and scalable enough to deliver significant improvements on challenging, large-scale data from a real-world production environment.

\begin{table*}[bp]
\centering
\caption{Performance of Coarse-Grained and Cross-Modal Retrieval Tasks on EIPM Dataset.}
\resizebox{2\columnwidth}{!}{
\begin{tabular}{lccccccccccc}
\toprule
      & \multicolumn{3}{c}{Image-to-Text Retrieval} &  & \multicolumn{3}{c}{Text-to-Image Retrieval} &  & \multicolumn{3}{c}{Coarse-grained Product Retrieval} \\ \cmidrule(lr){2-4} \cmidrule(lr){6-8} \cmidrule(l){10-12} 
Model & Recall@1  & Recall@5  & Recall@10 &  & Recall@1  & Recall@5  & Recall@10 &  & mAP@1   & mAP@5   & mAP@10   \\ \midrule
\multicolumn{12}{l}{\textit{BERT-based Baselines}}              \\
CLIP               &20.9  &42.1  &51.7  &  &20.5  &42.7  &52.8  &  &60.3  &62.8  &61.6  \\
ALBEF               &22.1  &43.1  &52.5  &  &21.8  &43.8  &53.7  &  &62.4  &65.5  &63.5  \\ 
BLIP              &23.7  &44.5  &53.6  &  &23.4  &45.1  &54.7  &  &63.1  &66.5  &64.7  \\ \midrule
\multicolumn{12}{l}{\textit{LLM-based Baselines}}           \\
LLM2CLIP           &27.5  &50.9  &60.3  &  &27.6  &52.0  &61.9  &  &67.4  &70.3  &68.7  \\
+ AGCL      &29.6  &52.7  &61.8  &  &29.5  &53.7  &63.2  &  &68.2  &71.2  &69.5  \\
VLM2Vec            &30.9  &52.2  &60.5  &  &35.8  &62.2  &71.6  &  &68.5  &72.6  &69.9  \\
+ AGCL        &33.2  &54.3  &62.1  &  &37.5  &63.7  &73.0  &  &69.4  &73.6  &70.7  \\ \bottomrule
\end{tabular}
}
\label{tab:main_online_res}
\end{table*}

\section{Details in training}
\label{sec:train_detail}

\begin{figure*}[!h]
    \centering
    \includegraphics[width=0.8\linewidth]{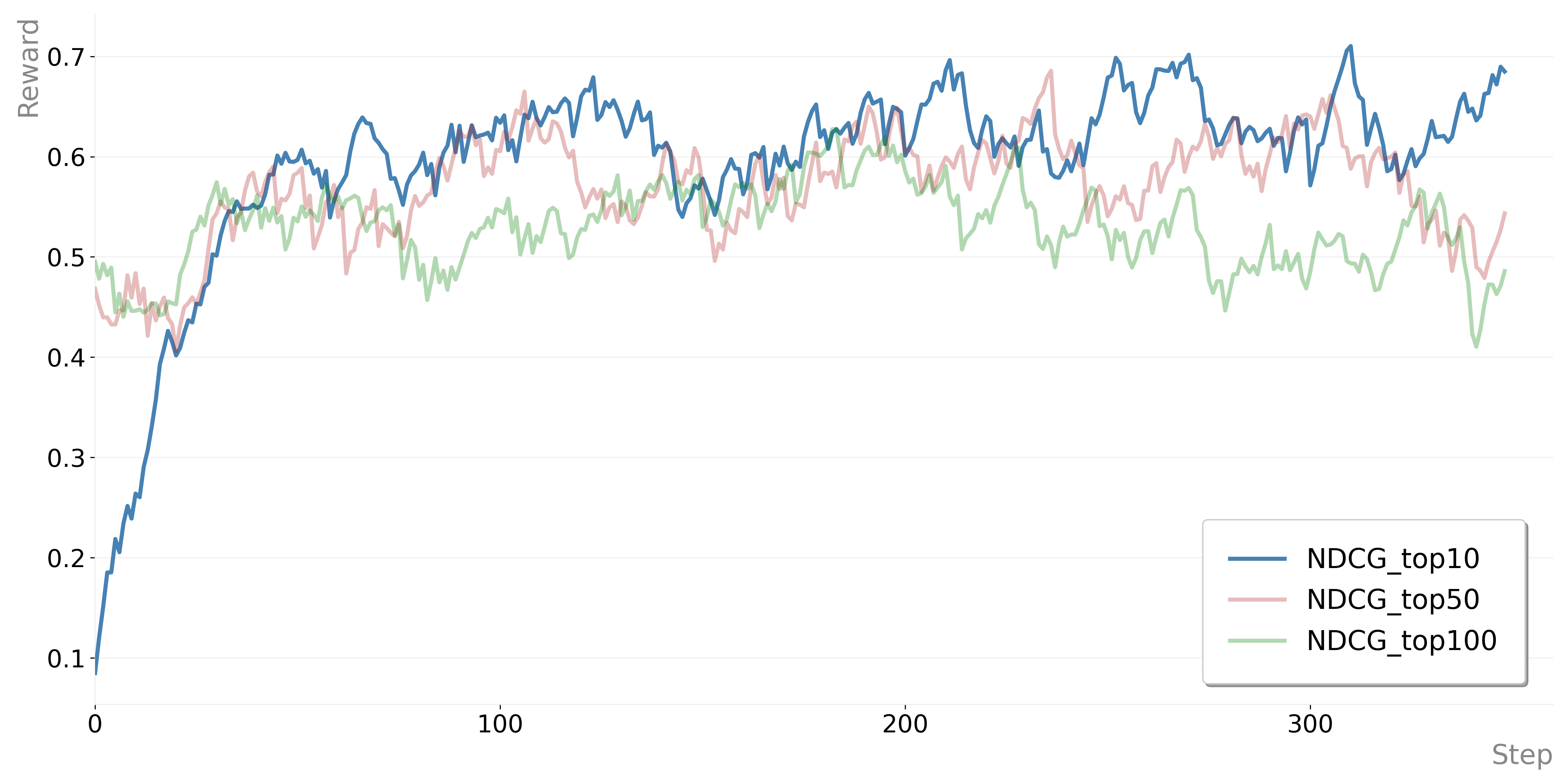}
    \caption{Training curves under different top‑k settings when using NDCG as the RL reward function.}
    \label{fig:ndcg_curve}
\end{figure*}

\begin{figure*}[!h]
    \centering
    \includegraphics[width=0.8\linewidth]{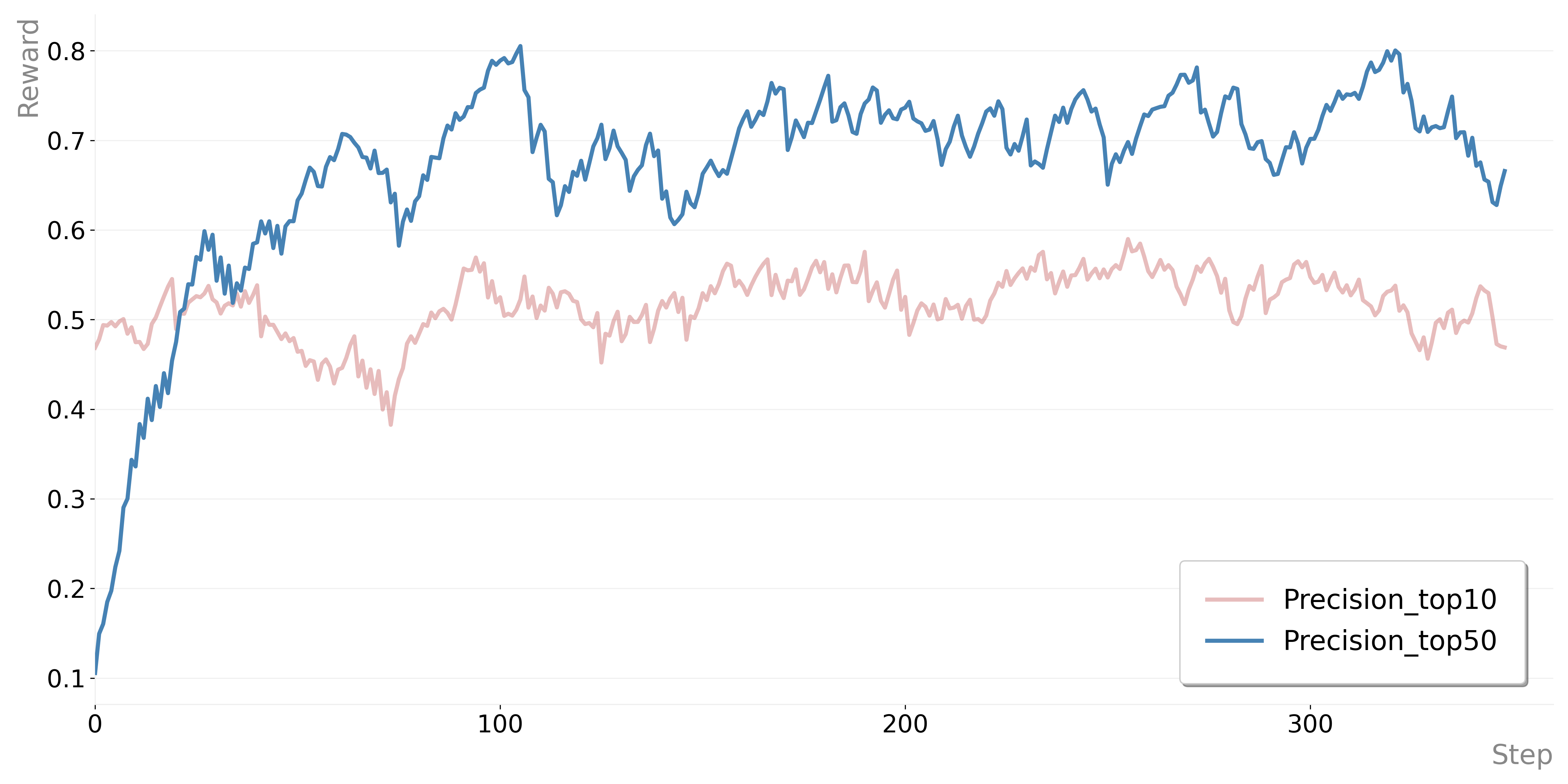}
    \caption{Training curves under different top‑k settings when using Precision as the RL reward function (training fails to converge when top‑k is set to 100).}
    \label{fig:precision_curve}
\end{figure*}

\begin{figure*}[!h]
    \centering
    \includegraphics[width=0.8\linewidth]{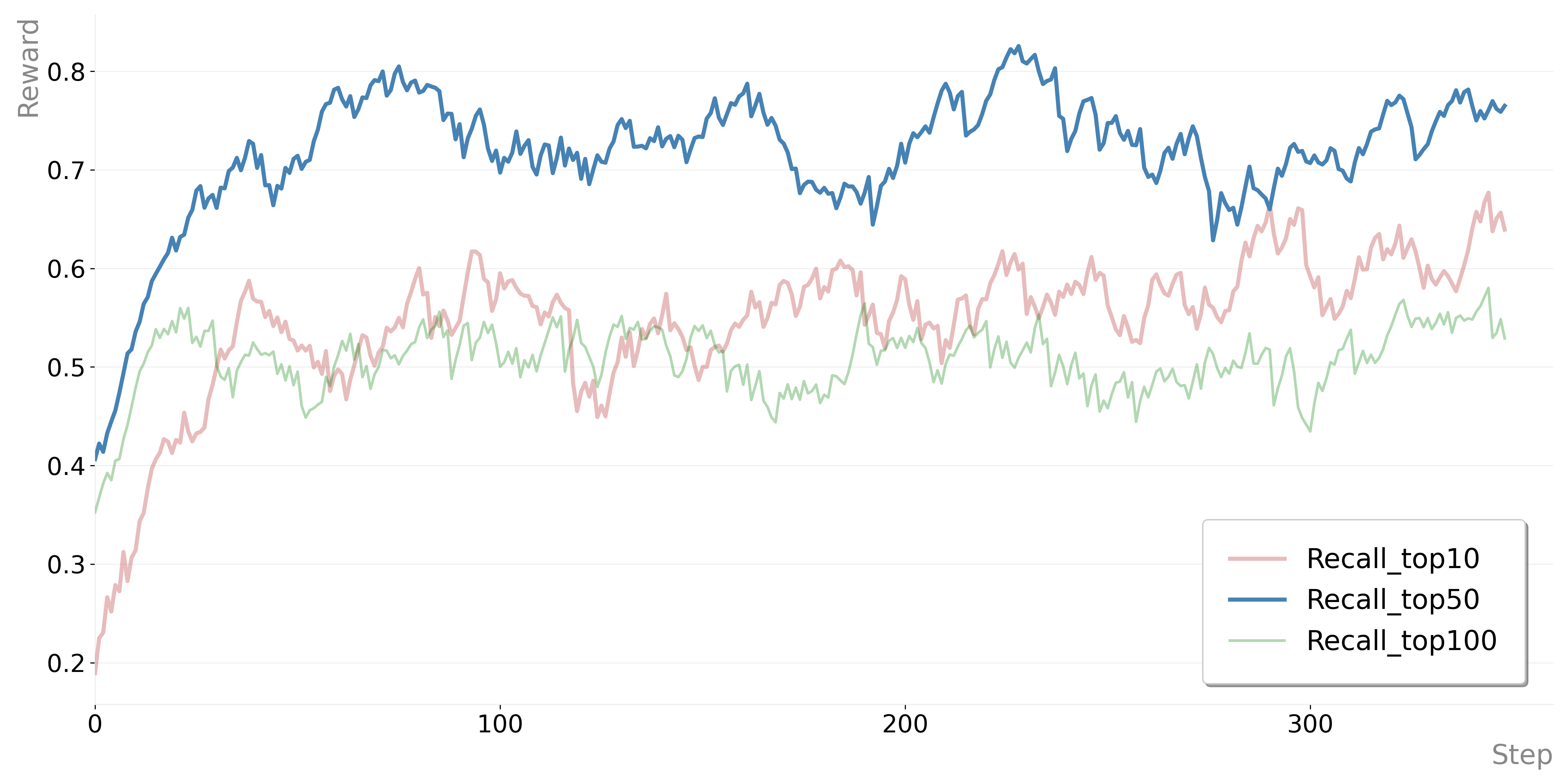}
    \caption{Training curves under different top‑k settings when using Recall as the RL reward function.}
    \label{fig:recall_curve}
\end{figure*}

We use Recall, Precision, and NDCG—three commonly used evaluation metrics for retrieval performance—as rewards, respectively, and compare the model convergence under different top-$k$ settings. As shown in Figures \ref{fig:ndcg_curve}-\ref{fig:recall_curve}, different metrics respond differently to the value of k. NDCG achieves optimal convergence at k=10; further increasing k weakens the ranking signals it relies on. In contrast, recall and precision peak at k=50, where the candidate pool provides dense learning signals without exhibiting saturation.

\section{Qualitative results}

\begin{figure*}[h]
    \centering
    \includegraphics[width=0.9\linewidth]{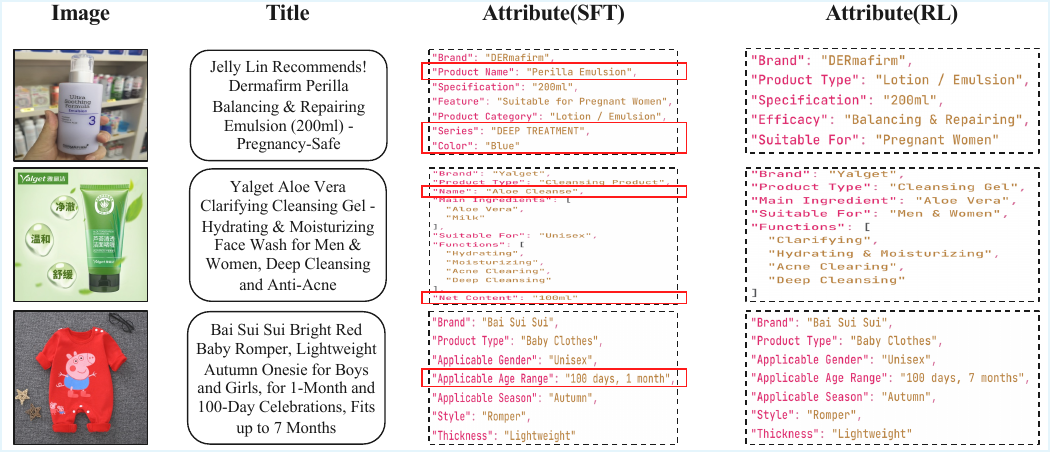}
    \caption{Cases of product item and generated attributes w/o RL.}
    \label{fig:attribute_case}
\end{figure*}

As illustrated in Figure \ref{fig:attribute_case}, our reinforcement learning (RL) framework significantly improves the quality of generated attributes. The baseline model often extracts noisy or overly general terms from verbose product titles. After optimization with our retrieval-aware RL policy, the model produces attributes that are substantially more succinct and accurate. The RL reward signal actively discourages the generation of terms that do not enhance discriminative power, compelling the model to focus only on the most essential and factually correct product features that are crucial for the fine-grained retrieval task.

\section{Cold start of the attribute generator}
\label{sec:generator}

We used a powerful model, Qwen-2.5-VL-72B-Instruct \cite{bai2025qwen2_5}, as an "attribute oracle" to generate precise attributes for a query, its positive target, and a hard negative. To enhance the quality and transparency of the generation, we prompted the oracle to first produce a Chain-of-Thought (CoT) reasoning process enclosed in \texttt{<think>} tags, followed by the final structured attributes enclosed in \texttt{<answer>} tags. This process forces the model to first "think" about the distinguishing features before extracting them.

We then measured the change in similarity scores produced by a base representation model, with and without the final oracle attributes from the \texttt{<answer>} block. As Figure \ref{fig:method_motivation}(b) illustrates, in the baseline condition, the hard negative achieves a similarity score perilously close to that of the positive target. However, when the model is augmented with the explicit key attributes, the representational space is reshaped, significantly widening the discriminative margin. This confirms our hypothesis: access to accurate, fine-grained attributes is critical for robustly distinguishing between similar items.

We argue that relying on such a massive oracle model is impractical for real-world deployment, and that extracting key product attributes is a relatively easy text generation task. Therefore, we adopt a knowledge distillation approach, transferring the oracle’s capability into a more inference-friendly model, Qwen-2.5-VL-3B-Instruct \cite{bai2025qwen2_5}. Specifically, we first perform Supervised Fine-tuning (SFT) on a dataset of 10,000 CoT examples generated by the oracle. This teaches our smaller generator model to mimic the oracle's reasoning process, generating both the \texttt{<think>} process and the final \texttt{<answer>} attributes.

\section{Preliminaries}

\subsection{Generative Model for Embedding Tasks}

Following the framework established by VLM2Vec \cite{jiang2025vlm2vec}, we adapt a generative model to the task of multimodal embedding. The goal is to train a model that can embed diverse data types into a shared semantic space. A training instance is represented as a query-positive target pair, ($q$, $t^+$), where both $q$ and $t^+$ can be an image, text, or an interleaved combination of both. To accommodate various downstream tasks, the query $q$ is conditioned with a specific task instruction (e.g., "Retrieve the most relevant product for this image").

The learning objective is to train a discriminative representation function $f(\cdot)$ using contrastive learning \cite{hadsell2006contra}. This objective aims to maximize the similarity score between the representations of a query and its positive target, $f(q)$ and $f(t^+)$, while simultaneously minimizing its similarity to all other negative targets $\{t^-\}$ in the batch. Given a Multimodal Large Language Model (MLLM) as our backbone, we obtain representations by extracting the hidden state of the final token from its last layer.

Formally, for a mini-batch of N pairs $\{(q_1, t_1),...,(q_N,t_N)\}$, the InfoNCE loss is defined as:
\begin{equation}
    \mathcal{L}=\frac{1}{N}\sum^N_{i=1}-log\frac{e^{s_{i,i}/\tau}}{e^{s_{i,i}/\tau}+\sum^N_{j\neq i}e^{s_{i,j}/\tau}}
    \label{infonce_loss},
\end{equation}
where $s_{i,j} = \text{cosine}(f(q_i), f(t_j))$ denotes the cosine similarity between the representations of query $q_i$ and target $t_j$. The function $f(\cdot)$ represents the MLLM-based embedding process, and $\tau$ is the temperature hyperparameter. Following VLM2Vec~\cite{jiang2025vlm2vec}, we set $\tau$ to 0.02.

\subsection{Promptable Multimodal Embeddings}

Traditional representation models generate a single, static embedding for each data item, which limits their flexibility for context-dependent tasks where a single item might be relevant to diverse intents (e.g., matching a product's brand versus its visual style). To address this, the paradigm of \textit{promptable embeddings} has emerged. This approach generates a context-aware representation by conditioning on a textual \textit{prompt}, transforming the embedding function to $f(\text{item, prompt})$ and allowing it to dynamically highlight task-relevant attributes  \cite{li2025highlighting}. The efficacy of this paradigm has been validated by models like VLM2Vec \cite{jiang2025vlm2vec} and E5-V \cite{jiang2024e5v} for creating domain-specific embeddings, and by architectures like GiVE \cite{li2024give} and FLAIR \cite{xiao2025flair} that enable finer-grained control via token-level text-image interactions. Building on this foundation, our work systematically studies promptable embeddings for retrieval targets and, crucially, develops strategies for their efficient, large-scale deployment.

\subsection{Large Batch Training with GradCache}

The efficacy of contrastive learning hinges on large batch sizes to ensure a diverse set of negative samples, yet the memory footprint of MLLMs makes this computationally prohibitive. 

To overcome this constraint, we employ GradCache \cite{gao2021gradcache}, a gradient caching technique that enables training with a very large effective batch size. The core idea is to decouple the backpropagation process of the contrastive loss from that of the encoder. Instead of performing a backward pass for the entire large batch, GradCache operates in two steps:
\begin{enumerate}
    \item Representation Gradient Computation and Caching: The large batch is divided into smaller mini-batches that fit into GPU memory. For each mini-batch, a forward pass is performed to compute embeddings, and these embeddings are used to calculate the representation-level gradients $\delta \mathcal{L}/\delta f(x_i)$ for all samples $x_i$ in the full batch. These small gradient tensors are then cached;
    \item Sub-batch Gradient Accumulation: For each mini-batch, a second forward and backward pass is performed. The full-batch gradients for the model parameters $\theta$ are then accumulated by multiplying the cached representation gradients with the local model gradients $\delta f(x_i)/\delta \theta$ and summing across all mini-batches.
\end{enumerate}

\end{document}